\documentclass[sigconf]{acmart}
\AtBeginDocument{%
  }

\usepackage{balance}
\usepackage{setspace} 
\usepackage{graphicx}
\usepackage{subcaption}
\usepackage{booktabs}
\usepackage{makecell}
\usepackage{multirow}
\usepackage{kotex}
\usepackage{listings}
\setcopyright{none}
\copyrightyear{2025}
\acmYear{2025}
\acmDOI{TBD}
\acmConference[IR-RAG @ SIGIR '25]{The Second Edition of the Workshop on Information Retrieval's Role in RAG Systems}{July 17, 2025}{Padua, Italy}
\acmISBN{978-1-4503-XXXX-X/2018/06}

\begin{document}

\title{HybridRAG: A Practical LLM-based ChatBot Framework based on Pre-Generated Q\&A over Raw Unstructured Documents}

\renewcommand{\shorttitle}{HybridRAG: A Practical LLM-based ChatBot Solution}

\author{Sungmoon Kim}
\affiliation{%
  \institution{Hanyang University}
  \city{Seoul}
  \country{South Korea}
}
\email{smk980510@hanyang.ac.kr}

\author{Hyuna Jeon}
\affiliation{%
  \institution{Hanyang University}
  \city{Seoul}
  \country{South Korea}
}
\email{younghyuna12@hanyang.ac.kr}

\author{Dahye Kim}
\affiliation{%
  \institution{Hanyang University}
  \city{Seoul}
  \country{South Korea}
}
\email{dahye99@hanyang.ac.kr}

\author{Mingyu Kim}
\affiliation{%
  \institution{Hanyang University}
  \city{Seoul}
  \country{South Korea}
}
\email{mingyu0519@hanyang.ac.kr}

\author{Dong-Kyu Chae}
\authornote{Corresponding authors.}
\affiliation{%
  \institution{Hanyang University}
  \city{Seoul}
  \country{South Korea}
}
\email{dongkyu@hanyang.ac.kr}

\author{Jiwoong Kim}
\authornotemark[1]
\affiliation{%
  \institution{Makebot Inc.}
  \city{Seoul}
  \country{South Korea}
}
\email{creative@makebot.ai}

\renewcommand{\shortauthors}{Kim et al.}

\begin{abstract}
\textit{Retrieval-Augmented Generation} (RAG) has emerged as a powerful approach for grounding Large Language Model (LLM)-based chatbot responses on external knowledge. However, existing RAG studies typically assume well-structured textual sources (e.g. Wikipedia or curated datasets) and perform retrieval and generation at query time, which can limit their applicability in real-world chatbot scenarios. In this paper, we present \textbf{HybridRAG}, a novel and practical RAG framework towards more accurate and faster chatbot responses. First, HybridRAG ingests raw, unstructured PDF documents containing complex layouts (text, tables, figures) via Optical Character Recognition (OCR) and layout analysis, and convert them into hierarchical text chunks. Then, it pre-generates a plausible question-answer (QA) knowledge base from the organized chunks using an LLM. At query time, user questions are matched against this QA bank to retrieve \textit{immediate answers} when possible, and only if no suitable QA match is found does our framework fall back to an on-the-fly response generation. Experiments on OHRBench demonstrate that our HybridRAG provides higher answer quality and lower latency compared to a standard RAG baseline. We believe that HybridRAG could be a practical solution for real-world chatbot applications that must handle large volumes of unstructured documents and lots of users under limited computational resources.
\end{abstract}

\maketitle

\section{Introduction}
\textit{Large Language Model} (LLM) based chatbots are increasingly being adopted to provide users with convenient, on-demand access to information and services \citep{chatbot_case1}. By interpreting user queries and generating precise answers, AI chatbots can operate 24/7 and reduce the need for manual information lookup or human customer service. A prominent strategy in recent chatbot systems is \textit{Retrieval-Augmented Generation} (RAG), which equips an LLM with an external knowledge base so that it can retrieve relevant context and ground its responses in up-to-date information \cite{chatbot_case2}. RAG has proven effective at reducing factual errors (hallucinations) by letting the model consult factual sources during generation, especially in domain-specific or knowledge-intensive tasks \cite{gao2023retrieval}.

Despite this progress, deploying RAG in real-world scenarios still faces important challenges. One challenge is that real-world documents are often unstructured and complex: In practice, enterprise knowledge bases include raw, scanned PDFs with figures or graphics, and forms with tables. Such content is not plain text and requires pre-processing like Optical Character Recognition (OCR) to be usable \cite{ohrbench}. However, most studies on RAG typically assume that external knowledge is available as clean, structured, textual data, typically curated from well-organized datasets such as Wikipedia or other textual corpora \cite{chatbot_case2}. Another challenge is that current RAG-based chatbots tend to be runtime-intensive: they perform retrieval and LLM inference for each user query. Such on-the-fly generation can lead to significant latency, especially problematic for enterprises operating under resource constraints and needing to serve high volumes of queries promptly. In this case, relying on the LLM alone to generate answers to numerous user queries would increase the response latency, ultimately leading to customer dissatisfaction and potential churn from the service.

In this work, we start with the following question: ``\textit{What if we could pre-generate plausible question-answer (QA) pairs from these raw document PDFs and utilize them at query time? And how much do they contribute to improving chatbot response quality and latency?}''
To answer this, we propose \textbf{HybridRAG}, a practical framework that enjoys the strengths of a pre-generated QA base to enhance the response time and quality. Our framework starts with an offline QA pre-generation from PDFs with an LLM. In order to convert a raw document (i.e., a scanned PDF of documents with non-selectable texts) into structured textual representations suitable for QA generation, we conduct MinerU-based layout analysis \cite{mineru} and OCR to extract text; for the detected non-textual elements such as tables and figures, we prompt an LLM (\textit{e.g.}, GPT4o-mini) to generate appropriate textual descriptions for them. Inspired by RAPTOR \cite{raptor}, we then apply a hierarchical chunking to organize the text representations into a tree of chunks ranging from fine-grained (paragraph or section) to coarse-grained (document summary). Finally, an LLM generates comprehensive and diverse set of plausible QA pairs from these hierarchical chunks. We design prompts with chain-of-thought reasoning and enforce that questions are answerable only using the chunk’s content, with no extraneous hallucination. In addition, we extract keywords to generate relevant QA pairs for each chunk node; here, we assign more keywords to higher layers of the hierarchy, which aggregate larger chunks of information, and fewer keywords to lower layers. Finally, these pre-generated QA pairs are indexed by their question embeddings.

At query time, HybridRAG first attempts to retrieve a matching question from the index and directly returns the corresponding answer if a close match is found (above a similarity threshold), without invoking the LLM generation. If no stored QA is a good match for the user’s query, HybridRAG uses the LLM to generate an answer on the fly based on the retrieved contexts. In this way, we expect that common or predictable questions can be efficiently and reliably answered, while still retaining the flexibility to handle unexpected queries.
 
To validate our HybridRAG, we conduct extensive experiments using OHRBench \cite{ohrbench}, consisting of 1,261 real-world unstructured PDF documents (in total 8,561 pages) spanning 7 domains (\textit{textbook}, \textit{law}, \textit{finance}, \textit{newspaper}, etc) and 8,498 ground truth QA pairs for benchmarking RAG systems. Our experiments show that HybridRAG achieves much lower average latency (and thus improved user experience) compared to a standard RAG pipeline. At the same time, HybridRAG achieves higher response quality than the baseline with the same LLM, demonstrating that the pre-generated knowledge base effectively aids answer correctness. %\textcolor{red}{The pre-generated QA pairs constructed by our approach cover a wide range of relevant questions (covering both detailed facts and high-level summaries) with faithful answers - 추후 실험결과에 따라 수정.}

%, 외부 Knowledge Base (KB)에 관련 문서를 저장하여 실시간으로 user query와 관련된 문서를 검색 및 답변에 활용하는 Retrieval Augmented Generation (RAG) 기반의 방식 
%\citep{facts_rag}
%을 사용하고 있다. 그러나 FAQ dataset을 만들기 위해서는 전문가의 지식이 필요하거나 관련 문서를 직접 탐색해야 하고, 충분한 양의 user query data가 있어야 가능하다. 또한 최신화가 필요할 때마다 새로운 인력과 fine-tuning을 위한 비용이 들어가게된다. RAG-based chatbot은 real-world (or raw) document를 이용하여 KB를 구축하기 어렵다는 점과 종종 잘못된 검색으로 인한 passage와 user query의 결합으로 부정확하거나 관련이 없는 답변을 한다.[RAG 관련해서 검색이 어려운 이유, or 기존 연구들은 이러이러하게 해결하려 했다.] 우리는 이러한 한계를 극복하기 위해 LLM을 이용해 자동으로 신뢰할 수 있는 real-world (or raw) document 기반의 Q\&A data를 생성함으로써 인력과 시간을 줄이고, 생성된 Q\&A data를 user query와 매칭하여 답변함으로써 빠르고 hallucination을 최소화하는 framework를 제안한다.\\

\section{Related Works}
\subsection{Unstructured Document Understanding}
Most previous RAG studies assume scenarios where refined textual data is directly available \citep{lewis2020retrieval, siriwardhana2021fine, misrahi2025adapting}, which may struggle to effectively handle questions on unstructured documents \citep{gao2023retrieval}. Several recent studies try to address such complex documents in the context of OCR \citep{blecher2023nougat, liu2024focus, wei2024general}, but simple OCR-based text extraction cannot sufficiently capture the characteristics of multimodal documents \cite{ohrbench}. Consequently, further research has aimed to improve document parsing performance \citep{chao2004layout, mineru, ohrbench}. More recently, several multimodal methods have been proposed to understand complex documents via integration with Vision-Language Models \citep{bai2025qwen2, chen2024expanding, chen2024far, wang2024qwen2}. Our HybridRAG is orthogonal to these approaches; they can be adopted to enhance the quality of our pre-generated QA pairs.
%대부분의 이전 RAG 연구들은 정제된 텍스트 데이터를 검색해오는 상황을 가정하고 있다 \textcolor{red}{[추가]}. 하지만 기존의 Text-Centric Document Retrieval은 실제 환경에서 접하게 되는 PDF, 이미지, 표와 같은 비정형 문서에 대한 질문은 잘 대처하지 못하는 한계가 존재한다\citep{gao2023retrieval}. 최근 RAG 기반 챗봇 시스템의 real world 에서의 적용 및 확장 가능성이 증가함에 따라, 보다 다양한 비정형 데이터를 포함하는 QA 시나리오를 다루기 위한 방식을 제안하고 있다. 대표적인 예로 이미지가 포함된 데이터에 대한 이해도를 높이기 위한 OCR 활용 연구\citep{blecher2023nougat, liu2024focus, wei2024general}가 진행되었지만, 단순 OCR기반 텍스트 추출은 데이터의 특성을 충분히 담지는 못한다\cite{ohrbench}. 이에 더 발전해서 Document Parsing을 고도화 하기 위한 연구\citep{chao2004layout, mineru, ohrbench}도 진행되어 왔다. 또한, 문서 내 표와 같은 반정형 정보나 이미지와 같은 비정형 정보를 모두 효과적으로 활용하기 위한 멀티모달 통합 방법들도 제안되고 있으며 \textcolor{red}{[추가]}, 이러한 접근은 Vision-Language 모델과 결합하여 비정형 정보가 포함된 QA 시나리오에 대한 적용 가능성을 높이고 있다.
%

\begin{figure}[t]
    \centering
    \includegraphics[width=\columnwidth]{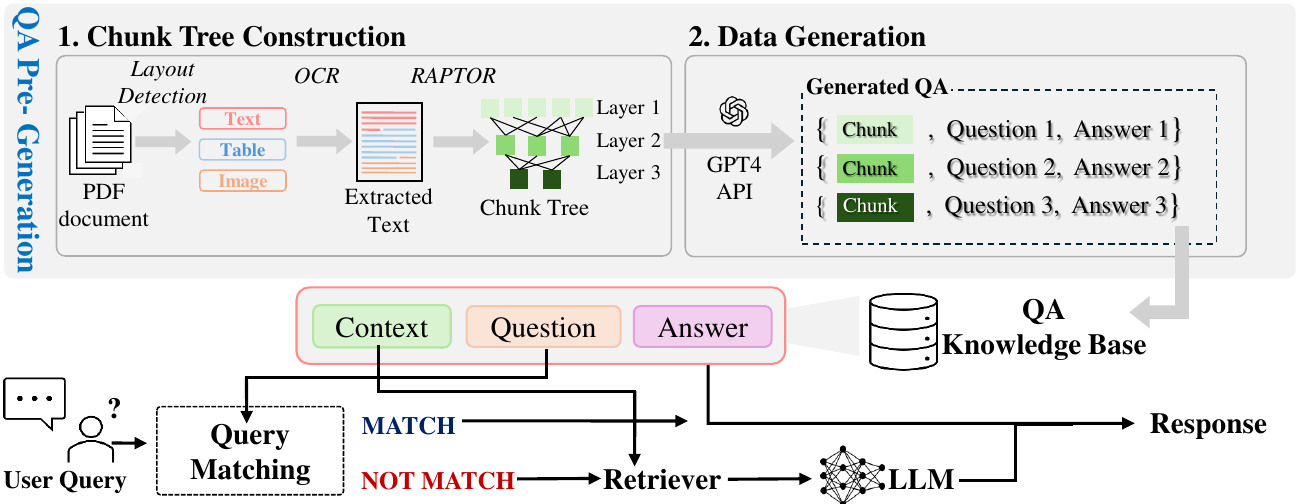}
   \vspace{-2mm}
   \caption{The overview of our HybridRAG.}
   \vspace{-2mm}
  \label{fig:fig1}
\end{figure}

\subsection{QA Generation}
Early studies on QA generation \citep{lewis2021paq, raffel2020exploring} have the limitation of focusing primarily on QA pairs including short, factoid-style answers.
Several methods \citep{nguyen2025urag, agrawal2024beyond} leverage a Frequently Asked Questions (FAQ) database to swiftly respond to frequent user queries, but the coverage of questions in such FAQ bases is very limited. Some work \citep{hr_faq} fine-tuned LLMs with an intensively curated QA dataset, but both QA data curation and fine-tuning LLMs may not be feasible under resource-limited situations. Different from the prior works, HybridRAG aims to generate numerous high-quality QA pairs via integrating layout analysis, OCR techniques, and hierarchical chunking to manage raw, structured PDF documents. 

%생성된 QA의 품질과 답변의 정확도를 보장하기는 어렵다는 한계점이 존재한다. 
%이에 반해, 우리는 다양한 형태의 비정형 문서로부터 LLM을 활용해 사전에 계층적인 고품질 QA 쌍을 생성하고, 이를 기반으로 오프라인 QA 베이스를 구성하는 방식을 채택한다.

% 기존-> 텍스트 기반. 이미지나 표에 대한 질문을 잘 대처하지 못함
% 기존-> 생성 모듈에서 사용되는 LLM의 성능에 영향을 크게 받음. 소형 모델은 잘 동작하지 못함 

\section{Method}
The proposed framework consists of an offline QA pre-generation phase and an online query-time phase. Figure \ref{fig:fig1} illustrates the overview of our framework.%QA Generation 단계에서는 주어진 문서들을 분석해서 QA 쌍을 생성하고, query time에서는 사용자 질의에 대해 적절한 답변을 제공하기 위해 생성된 QA base를 활용한다.

\subsection{QA Pre-Generation}\label{qag}
%QA Generation Phase에서는 입력된 PDF 문서로부터 검색 가능한 Knowledge Base(KB)를 구축한다. 문서는 레이아웃 분석과 OCR을 통해 텍스트로 변환되고, RAPTOR기반의 계층적 chunking을 통해 트리 구조를 만든다. 각 청크로부터 생성된 QA 쌍은 임베딩 유사도 기반 필터링을 거쳐 KB에 저장된다. 
%The QA pre-generation phase aims to construct QA pairs from input PDF documents. These documents are converted into text through layout analysis and OCR, and then structured into a tree using RAPTOR-based hierarchical chunking. 

\subsubsection{Document Preprocessing}\label{pp}
%본 단계에서는 PDF문서를 입력받아 이를 위해 우선 pdf2image 라이브러리로 PDF를 페이지 단위 이미지로 변환하고, PaddleOCR \textcolor{red}{[추가]} 로 텍스트 블록을 인식한다. 표나 그림과 같이 OCR만으로 정확한 정보를 추출하기 어려운 테이블, 그림 등의 요소는 Gpt-4o 모델을 통해 아래의 프롬프트를 줘서 디스크립션을 생성한다:  
Each input PDF document is first processed by the MinerU \cite{mineru}, which segments each page into layouts of type text, table, and image, and outputs their bounding box coordinates in a machine-readable format (e.g., Markdown or JSON). Text blocks within the detected regions are then extracted using PaddleOCR\footnote{https://github.com/PaddlePaddle/PaddleOCR}. Non‑textual layouts (tables and figures) are converted to textual descriptions via a dedicated LLM. Here, we use GPT-4o with the following prompt to generate descriptive text:

\begin{center}
%\textbf{Prompt used for image description}
%\vspace{0.5em}

\begin{minipage}{0.95\linewidth}
\begin{spacing}{0.9}   % 기본 줄간격의 90%로 축소
      \itshape
You are an image description expert who accurately outputs chunk text based on the information contained in the image. You must analyze the content without omission, avoid duplication or misinterpretation, and write the result as a coherent paragraph without any markdown formatting or external commentary.
\end{spacing}
\end{minipage}
\vspace{-2mm}
\end{center}
%\vspace{-1.3em}
%이렇게 유형별로 생성된 텍스트를 하나로 모아 후속 단계에서 활용할 수 있는 형태로 구조화하며, 이는 이후 단계에서 사용할 수 있는 청크이다. 마지막으로 문서에 포함된 여러 요소가 전체 문맥에서 유기적으로 연결되도록 페이지, 좌표, 순서 등의 메타데이터를 함께 저장한다. 
The extracted texts via OCR and generated ones with GPT-4o are then consolidated and structured into chunks for later use. Finally, to maintain contextual coherence among various document elements, we also store their metadata such as page number, coordinates, and sequential order.

\subsubsection{Hierarchical Chunking}
Inspired by RAPTOR \cite{raptor}, we partitioned the entire document chunks into a hierarchical tree structure; we place compressed and summarized information about the whole document at the top-level node, and progressively include more specific and detailed content in lower-level nodes. Structuring the document into a tree in this way allows effective and flexible responses to user queries, ranging from general context to specific details, thereby enabling comprehensiveness QA generation.
%Inspired by RAPTOR \cite{raptor}, 우리는 문서 전체를 계층적 트리 구조로 분할하였다; 최상위 노드에는 문서 전반을 압축·요약한 정보를 배치하고, 하위 노드로 갈수록 보다 구체적이고 세부적인 내용을 포함하도록 했다. 이와 같이 트리 형태로 구조화하면, user query에 대해 문서의 전체적인 맥락부터 특정 세부 사항까지 효과적으로 유연하게 답변할 수 있어 comprehensiveness와 diversity를 동시에 고려한 고품질의 응답 생성이 가능하다. 

\subsubsection{Keyword Extraction}
%각 노드와 관련성 높은 QA쌍을 생성하기 위해 Gpt-4o-mini을 활용하여 핵심 키워드를 추출한다. RAPTOR 트리의 각 레이어 깊이에 따라 정보량이 다르다는 특성을 반영하여 여러 청크의 내용이 집약되어있는 상위 레이어일 수록 많은 수를, 하위 레이어일수록 적은 수의 키워드를 맵핑하여 adaptive하게 추출한다.
To generate as relevant QA pairs for each node as possible, we extract core keywords for each node via GPT-4o-mini. Reflecting the characteristic of the chunk tree structure (higher layers aggregate content from multiple chunks and thus contain richer information), we map a larger number of keywords to upper layers and fewer keywords to lower layers.

\subsubsection{QA Generation}
%실제 사용자가 질문할법한 QA 쌍을 생성하기 위해 Chain-of-Thought 기반 프롬프트를 적용하였다. LLM (gpt-4o-mini)이 5단계의 체계적인 사고과정을 거치도록 설계하였으며, 텍스트 내 명시된 정보만을 기반으로 질문을 생성하도록 설계하였다. 또한 기존 생성된 QA와의 중복을 방지하는 제약조건을 함께 부여하였다. QA 품질 향상을 위해 프롬프트에 원본 텍스트, 주어진 키워드, 구체적인 사고과정 설명, 생성된 QA 쌍으로 구성된 구체적인 예시를 포함하였다. 또한 청크별로 생성되는 QA의 수는 해당 청크에서 추출된 키워드 수와 동일하게 설정하여 청크 내 다양한 핵심정보를 누락 없이 반영할 수 있도록 하였다. 구체적인 프롬프트는 다음과 같다:
To generate QA pairs reflective of realistic user queries, we apply a chain-of-thought-based prompting approach. The prompt is designed to guide the LLM (GPT-4o-mini) through a structured, five-step reasoning process in order to ensure that questions were generated exclusively from explicitly stated information in the text. Additionally, constraints were included to prevent redundancy with previously generated QA pairs. To enhance QA quality, the prompt contained detailed examples comprising the original text, provided keywords, explicit reasoning steps, and resulting QA pairs. Furthermore, the number of QA pairs generated per chunk was set equal to the number of keywords extracted from that chunk, ensuring comprehensive coverage of diverse key information. The detailed prompt is as follows:

\begin{center}
\textbf{<Prompt used for QA generation>}\\[0.5em]

\begin{minipage}{0.95\linewidth}
\begin{spacing}{0.9}   % 기본 줄간격의 90%로 축소
      \itshape
You are an AI specialized in generating QAs from documents. Your mission is to analyze the document, follow the instructions, and generate RAG-style question-answer pairs based on the document.
    RAG-style refers to a question that needs to be answered by retrieving relevant context from an external document based on the question, so the question MUST obey the following criteria:
\end{spacing} 
\end{minipage}

\begin{minipage}{0.95\linewidth}
\begin{spacing}{0.9}   % 기본 줄간격의 90%로 축소
      \itshape
    1. Question should represent a plausible inquiry that a person (who has not seen the page) might ask about the information uniquely presented on this page. The questions should not reference this specific page directly (by page number, pointing to a specific paragraph or figure, and never refer to the document using phrases like ’in the document’), nor should they quote the text verbatim. They should use natural language reflecting how someone might inquire about the page’s content without direct access. \\
    2. Question must contain all information and context/background necessary to answer without the document. Do not include phrases like "according to the document" in the question.\\
    3. Question must not contain any ambiguous references, such as 'he', 'she', 'it', 'the report', 'the paper', and 'the document'. You MUST use their complete names.
\end{spacing}
\end{minipage}
\end{center}

\vspace{-3mm}
\begin{center}
\textbf{<System prompt used for QA generation>}\\[0.5em]

\begin{minipage}{0.95\linewidth}
\begin{spacing}{0.9}   % 기본 줄간격의 90%로 축소
      \itshape
- Instruction:\\
1. Analyze the text above and the given keywords.\\
2. Create new, meaningful question–answer pairs that a user might naturally ask about this text.\\
3. Do not duplicate any previously generated Q\&A.\\
4. Only generate questions that can be answered explicitly by the text.\\
5. Provide concise, direct answers without extra elaboration.\\

The answer form should be as diverse as possible, including [Yes/No, Numeric, String, List].\\

- Output format:\\
Question:\\
Answer:\\
\end{spacing}
\end{minipage}
\end{center}

\iffalse
\begin{center}
\textbf{few-shot example}\\[0.5em]

\begin{minipage}{0.95\linewidth}
\itshape
    \#\#\# text:\\ "With direct access to human-written reference as memory, retrieval-augmented generation has achieved much progress in various text generation tasks. In this framework, better memory typically leads to better generation (the 'primal problem'). The proposed Selfmem approach leverages an iterative memory to improve performance."\\
    \#\#\# keywords: ["retrieval-augmented generation", "primal problem", "Selfmem"]\\
    (Reasoning steps)\\
    1. Identify main content: retrieval-augmented approach, concept of 'better memory => better generation'\\
    2. Form a question about what tasks or idea the snippet highlights\\
    3. The answer must be directly found in the snippet\\
    Question: What core idea does the Selfmem framework build upon?\\
    Answer: It uses an iterative memory mechanism, where better generation leads to better memory, improving overall performance.\\
\end{minipage}
\end{center}
\fi

\vspace{-6mm}
\subsection{HybridRAG}
We embed the questions from the pre-generated QAs using the BGE-M3 (dense retrieval) \cite{bgem3}. At query time, a given user query is embedded using the same model, and similarity scores are calculated via the inner product between embeddings. The top-3 QA pairs are then retrieved, and if the highest similarity score exceeds a predefined threshold (e.g., 0.9), the answer from the most similar QA pair is directly returned. However, if the similarity score is below the threshold, the chunks corresponding to the retrieved top-3 QA pairs are aggregated and provided along with the user query as input to an LLM to generate the final response. We used Llama3.2-3B-Instruct \cite{llama3} and Qwen2.5-3B-Instruct \cite{qwen25} as the generative LLMs for generating on-the-fly responses.

%먼저 사전 생성된 CQA 쌍에서 Question만을 임베딩한다. 이때 임베딩모델로는 BGE-M3 \textcolor{red}{[추가]} 를 활용한다. 사용자 질의가 들어오면 이를 동일 모델로 임베딩한 후, 이를 사전 계산된 question 임베딩과의 내적을 통해 유사도를 계산한다. 유사도를 비교하여 사용자 질의와 가장 유사한 컨텍스트를 포함하는 상위 k개의 QA 쌍을 검색하며, 이때 유사도 점수가 사전 설정된 임계값(0.9)을 초과할 경우, 가장 유사한 QA 쌍의 답변(answer)을 바로 반환한다. 반면 유사도 점수가 임계값 이하인 경우에는, 검색된 상위 k개의 QA 쌍에 대응되는 chunk들을 통합한 후  이를 사용자 질의와 함께 생성형 LLM의 입력으로 제공하여 최종 답변을 생성한다.

\iffalse
\begin{center}
\textbf{User Prompt used for RAG}\\[0.5em]

\begin{minipage}{0.95\linewidth}
\itshape
 You are an expert, you have been provided with a question and documents retrieved based on that question. Your task is to search the content and answer these questions using both the retrieved information. \\
You **MUST** answer the questions briefly with one or two words or very short sentences, devoid of additional elaborations.\\
Write the answers within <response></response>. If you cannot find answer from retrieved Documents, say: "Not answerable".\\
\end{minipage}
\end{center}
\fi

\section{Experiments}

\subsection{Experimental Settings}
\subsubsection{Benchmark}
We employed OHRBench curated by the authors of \cite{ohrbench}, including 1,261 PDFs spanning 8,561 pages. These PDFs were obtained from seven real-world RAG application domains, including Finance, Administration, Law, etc. Furthermore, this benchmark provides a diverse set of ground truth QA pairs derived from multimodal elements in those documents. Table \ref{tab:dataset} shows the statistics of OHRBench.

\begin{table}[ht]
  \centering
  \caption{Benchmark statistics.}
  \vspace{-4mm}
  \label{tab:dataset}
  \small
  \begin{tabular}{lrrc}
    \toprule
    Domains       & \# PDFs  & \# Pages & \# Q\&A pairs \\
    \midrule
    Law            &  95  & 1,187  & 1,142            \\
    Finance        &  65  & 2,133  & 1,367            \\
    Textbook       & 504  &  678  & 1,125           \\
    Manual         &  87  & 1,724  & 1,151            \\
    Newspaper      & 279  &  487  & 1,202              \\
    Academic       &  85  & 1,011  & 1,179            \\
    Administration & 146  & 1,341  & 1,332            \\
    \midrule
    Total          & 1,261 & 8,561  & 8,498            \\
    \bottomrule
  \end{tabular}
  \vspace{-2mm}
\end{table}

\begin{table*}[t]
  \caption{Performance comparison w.r.t. response quality and latency. Best performance is in \textbf{bold}; second-best is underlined.}
  \vspace{-4mm}
  \centering
  \footnotesize
  \begin{tabular}{ll|cccc|cccc|cccc}
    \toprule
    & & \multicolumn{4}{c}{Standard RAG}
      & \multicolumn{4}{c}{HybridRAG (simplified)}
      & \multicolumn{4}{c}{HybridRAG (Ours)} \\
    \cmidrule(lr){3-6} \cmidrule(lr){7-10} \cmidrule(lr){11-14}
    LLM        & Domain        & F1($\uparrow$)    & BERT($\uparrow$) & ROUGE($\uparrow$) & Latency($\downarrow$)
               & F1($\uparrow$)    & BERT($\uparrow$) & ROUGE($\uparrow$) & Latency($\downarrow$)
               & F1($\uparrow$)    & BERT($\uparrow$) & ROUGE($\uparrow$) & Latency($\downarrow$) \\
    \midrule
    \multirow{8}{*}{Llama3.2}
      & Academic       & 21.27 & 0.7521 & 0.2671 & 0.899s
                      & 20.10 & 0.7442 & 0.2510 & 0.699s 
                      & 20.87 & 0.7417 & 0.2459 & 0.843s \\
      & Administ. & 25.09 & 0.7678 & 0.2894 & 1.046s
                      & 24.56 & 0.7608 & 0.2774 & 0.747s
                      & 26.31 & 0.7640 & 0.2954 & 0.842s \\
      & Finance        & 15.26 & 0.7133 & 0.1795 & 1.185s
                      & 14.82 & 0.7106 & 0.1780 & 0.774s
                      & 17.51 & 0.7134 & 0.2034 & 1.085s \\
      & Law            & 31.62 & 0.7805 & 0.3572 & 0.889s
                      & 30.99 & 0.7776 & 0.3370 & 0.572s
                      & 31.14 & 0.7793 & 0.3471 & 0.562s \\
      & Manual         & 31.07 & 0.7896 & 0.3476 & 0.996s
                      & 27.82 & 0.7761 & 0.3135 & 0.694s
                      & 28.48 & 0.7747 & 0.3201 & 0.985s \\
      & Newspaper      & 20.91 & 0.8126 & 0.1938 & 5.021s
                      & 27.59 & 0.8346 & 0.2201 & 1.236s
                      & 26.09 & 0.8310 & 0.2108 & 1.338s \\
      & Textbook       & 17.50 & 0.7561 & 0.2662 & 1.760s
                      & 19.34 & 0.7692 & 0.2825 & 0.772s
                      & 20.12 & 0.7527 & 0.2596 & 0.866s \\
                      \cmidrule(lr){2-14}
      & Avg.           & 23.25 & 0.7674 & \textbf{0.2715} & 1.685s
                      & \underline{23.60} & \underline{0.7676} & 0.2656 & \textbf{0.785s}
                      & \textbf{24.36} & \textbf{0.7692} & \underline{0.2689} & \underline{0.931s} \\
    \midrule
    \multirow{8}{*}{Qwen2.5}
      & Academic       & 14.14 & 0.7069 & 0.1704 & 0.403s
                      & 16.10 & 0.7137 & 0.1887 & 0.362s
                      & 16.85 & 0.7127 & 0.1894 & 0.390s \\
      & Administ. & 14.86 & 0.7054 & 0.1723 & 0.409s
                      & 18.25 & 0.7203 & 0.2049 & 0.372s
                      & 21.80 & 0.7307 & 0.2359 & 0.412s \\
      & Finance        &  7.71 & 0.6584 & 0.0816 & 0.331s
                      & 10.08 & 0.6748 & 0.1174 & 0.426s
                      & 12.31 & 0.6826 & 0.1344 & 0.403s \\
      & Law            & 19.42 & 0.7394 & 0.1985 & 0.325s
                      & 23.03 & 0.7490 & 0.2415 & 0.287s
                      & 25.34 & 0.7553 & 0.2628 & 0.315s \\
      & Manual         & 21.53 & 0.7396 & 0.2295 & 0.396s
                      & 22.80 & 0.7438 & 0.2448 & 0.318s
                      & 23.15 & 0.7462 & 0.2526 & 0.372s \\
      & Newspaper      & 17.25 & 0.7133 & 0.1288 & 0.490s
                      & 25.22 & 0.7658 & 0.1844 & 0.495s
                      & 22.78 & 0.7557 & 0.1728 & 0.455s \\
      & Textbook       & 12.37 & 0.6947 & 0.1571 & 0.458s
                      & 15.96 & 0.7114 & 0.1957 & 0.354s
                      & 13.70 & 0.6934 & 0.1395 & 0.295s \\
                      \cmidrule(lr){2-14}
      & Avg.           & 15.33 & 0.7082 & 0.1626 & 0.402s
                      & \underline{18.78} & \textbf{0.7255} & \underline{0.1968} & \textbf{0.373s}
                      & \textbf{19.42} & \underline{0.7252} & \textbf{0.1982} & \underline{0.377s} \\
    \bottomrule
  \end{tabular}
  \vspace{-1mm}
  \label{tab:main}
\end{table*}

\subsubsection{Metrics}
In terms of response quality, we measure how well the answers generated by a RAG system match the ground truth answers provided in OHRBench (the QA pairs), using the following three metrics: \textbf{F1-score} measures the overlap between the model response and the ground truth answer based on word tokens; \textbf{ROUGE-L} evaluates the Longest Common Subsequence (LCS) between the model's answer and the ground truth answer; and \textbf{BERTScore} utilizes contextual embeddings from BERT to measure semantic similarity between the generated answer and the ground truth answer. For response \textbf{latency}, we measure the average time interval from receiving a user query to completing the response. We used a server equipped with an NVIDIA RTX 3090 GPU.

%RAG system의 latency를 측정하기 위해, 우리는 검색 단계부터 최종 답변 생성까지 걸리는 end-to-end latency를 측정했다.
% For measuring latency, we use \textbf{TTFT} (Time-to-First-Token): the time interval from receiving a user query to generating the first token of the chatbot's response. 

\subsubsection{Baselines}
In order to highlight the effectiveness of our method, we implement two baselines: (1) \textbf{Standard RAG} follows the original RAG pipeline without any pre-generated QAs. It only adopts MinerU \cite{mineru}-based OCR to recognize texts in the given PDFs, without considering figures or tables. The chunk size is set to 1024 tokens without overlap, following \cite{ohrbench}. Top-3 most similar text chunks are retrieved to aid LLM response generation. (2) \textbf{Simplified HybridRAG} involves pre-generating QA pairs, but it only considers plain text extracted by MinerU-based OCR (identical to the Standard RAG setup). Other aspects, such as embedding models or LLMs, are identical across all three compared models.
%우리는 Standard RAG와 HybridRAG의 비교 실험을 위해 retriever로 BGE-M3 \cite{bgem3}의 2가지 retriever를 사용한다. (1) 문장 단위의 벡터를 통해 의미 검색을 하는 dense retirever, (2) token 단위의 중요도 weight를 통한 검색을 하는 sparse retriever를  사용한다. 

%Standard RAG는 MinerU로 추출된 text를 knowledgebase로 사용했으며,  chunk size는 \cite{ohrbench}의 setting과 동일하게 overlap 없이 1024로 설정했다. query와 가장 유사한 top-3개의 chunk를 검색하고 LLM에게 제공하여 답변을 생성했다.

%HybridRAG는 preprocessing 유무에 따라 2가지 버전으로 실험을 했다. preprocessing이 없는 simple version (baseline)은 Section \ref{pp}에서 언급된 방법을 사용하지 않고 MinerU로 추출된 text에서 생성된 QA pairs를 knowledgebase로 사용했고, HybridRAG (Ours)는 Section \ref{qag}의 모든 방법을 충실히 따른 version이다. 

\subsection{Response Quality and Latency}

Table \ref{tab:main} compares the performance of a standard RAG, HybridRAG and its simplified version in terms of response quality and latency. Overall, the simplified HybridRAG outperforms Standard RAG, which demonstrates that the pre‑generated QA pairs (even though they were based solely on texts in the documents) have a positive effect on both latency and response quality. With Llama3.2, latency is especially improved, and with Qwen2.5, there is a noticeable answer quality gain. In addition, HybridRAG (Ours)—which incorporates layout detection, OCR, LLM‑driven table/figure description, and hierarchical chunking—further boosts answer quality over Simplified HybridRAG. Especially, the domains with a high proportion of chart/table–based queries in OHRBench (Administration: 562, Finance: 1,038 QA pairs) show the largest F1 gains from our HybridRAG: The Administration domain improves by 19\% with Qwen2.5, and Finance by 22\% compared to Simplified HybridRAG. 

%This confirms that our strategy to ingest multimodal elements in raw PDFs highly contribute to generating high-quality .

%We can observe that our HybridRAG consistently outperforms the two compared methods across all the metrics in terms of response quality. Specifically, our HybridRAG method demonstrates higher F1 scores. 
%HybridRAG의 2가지 version (baseline and ours) 모두 Standard RAG에 비해 평균적으로 더 나은 답변 성능을 보인다. 이를 통해 생성된 QA pairs를 knowledgebase로 사용하는 것에 대한 효과를 알 수 다. 
%뿐만 아니라 the end-to-end latency is significantly reduced by our method
%, particularly notable in domains such as Administration, Law, Newspaper and Finance, highlighting the efficiency and practicality of HybridRAG for real-world, resource-constrained deployments.
%OHRBench가 제공하는 ground truth QA pairs 중 chart and table에 대한 query 비율이 높은 domain은 academic (605개), administration (562개), finance (1,038개)이며, HybridRAG (Ours)의 F1-Score가 HybridRAG (baseline)에 비해 크게 증가하는 것을 볼 수 있다. 특히 Qwen-3B-Instruct에서 두드러지게 나타나는데, academic에서는 5\%, administration에서는 19\%, finance에서는 22\%까지 증가한다. 이를 통해 text뿐만 아니라 image나 table 등 다양한 요소들이 포함된 raw PDF를 HybridRAG (Ours)가 효과적으로 처리할 수 있음을 (간접적으로?) 확인할 수 있다.
% 2가지 HybridRAG 모두 Standard RAG의 답변 품질보다 좋음. 이를 통해 QA pair를 KB로 사용하는 것의 이점~~  
%Additionally, the end-to-end latency is significantly reduced by our method, particularly notable in domains such as Administration, Law, and Finance, highlighting the efficiency and practicality of HybridRAG for real-world, resource-constrained deployments.

\subsection{Quality of Pre-generated QAs}
% OHRBench, 4o-mini, pp+mini의 G-Eval 실험 과정 및 결과 
This section aims to evaluate the quality of our pre-generated QA pairs. Since there is no human-crafted references or ground truth, we adopt the LLM-judge approach, specifically G-Eval \cite{geval}. We assess the quality of QA pairs across the following four perspectives proposed by QGEval \cite{qgeval}: \textbf{Context-Question-Answer Relevance (CQAR)} assesses how well the generated question and answer align contextually with the source document; \textbf{Answerability} measures the extent to which the generated question is answerable given the provided context; \textbf{Clarity} evaluates the preciseness and unambiguity of the generated QAs; \textbf{Fluency} reflects the grammatical correctness and naturalness of the generated text.
For comparison, the QA pairs provided by OHRBench, which were oriented from the same PDF sources with ours, are also evaluated in the same way. Approximately 5,000 QA pairs were sampled from each set and evaluated via G-Eval.
\vspace{-1mm}
\begin{figure}[h]
    \centering
    \includegraphics[width=0.8\columnwidth]{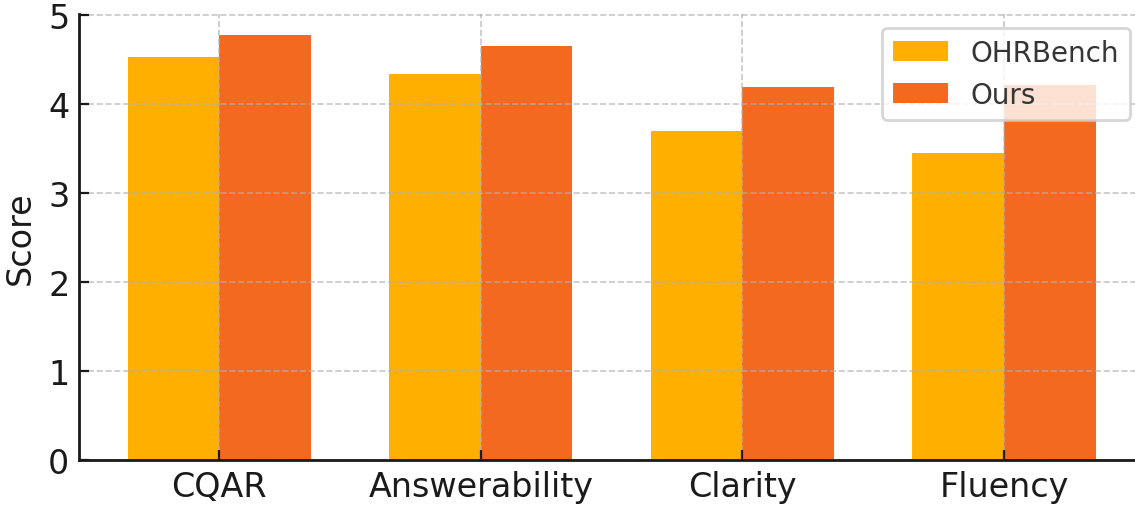}
   \vspace{-1mm}
   \caption{Quality evaluation of the pre-generated QAs.}
   \vspace{-1mm}
  \label{fig:QGEval}
\end{figure}

Figure \ref{fig:QGEval} shows the results. Compared to the QA pairs provided by OHRBench, those generated by our method demonstrated superior quality w.r.t. all four dimensions: CQAR (+0.24), answerability (+0.31), clarity (+0.50), and fluency (+0.76).

% 생성된 QA pairs의 다양성을 측정하기 위해 self-BLUE를 이용. self-BELU에 대한 간략한 설명, 도메인별 결과

\subsection{Threshold Analysis}
\begin{figure}[htbp]
  \centering
  \begin{subfigure}[b]{0.49\linewidth}
    \centering
    \includegraphics[width=\linewidth]{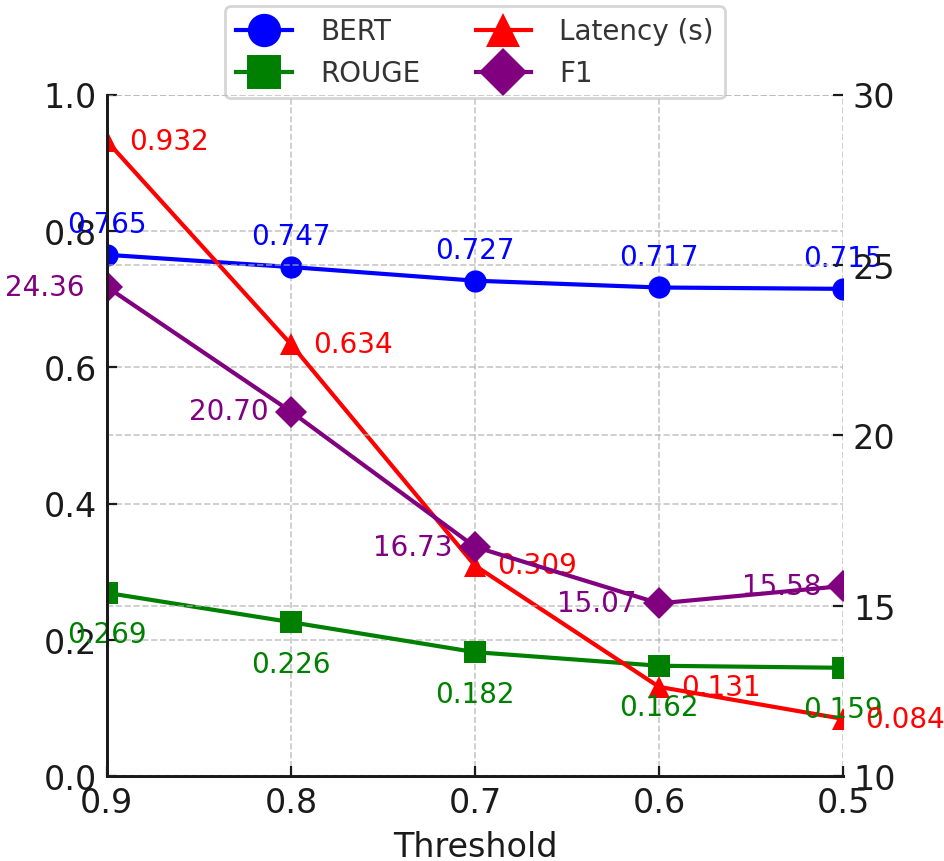}
    \caption{Llama3-3B}
    \label{fig:plot_old}
  \end{subfigure}
  %\quad
  \begin{subfigure}[b]{0.49\linewidth}
    \centering
    \includegraphics[width=\linewidth]{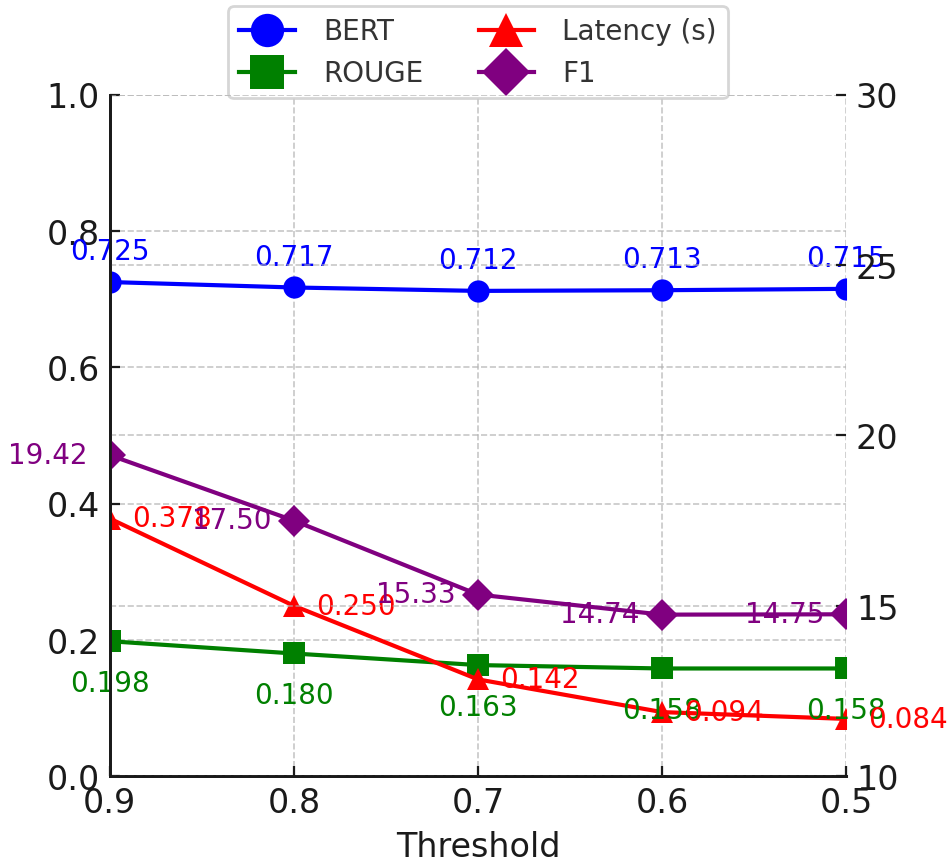}
    \caption{QWEN2.5-3B}
    \label{fig:plot_new}
  \end{subfigure}
  \vspace{-1mm}
  \caption{Performance depending on different threshold.}
   \vspace{-1mm}
  \label{fig:threshold}
\end{figure}

Lowering the threshold increases the proportion of responses directly returned from the pre-generated QA base. Specifically, at a threshold of 0.9, stored responses handle 13\% of the total test queries; however, at a threshold of 0.7, this proportion rises significantly to 80\%. Consequently, as shown in Figure \ref{fig:threshold}, while this reduces the average latency further, it introduces a trade-off by simultaneously decreasing the quality of responses.
%에서, 매칭 threshold가 낮을수록 응답을 그대로 반환하는 비율이 증가하여 평균 end-to-end latency는 감소하지만, 그에 따라 답변 품질은 저하되는 trade-off를 볼 수 있다. 이러한 결과를 통해 대량의 user query를 처리 해야 하는 기업에서, resource 상황에 맞는 적절한 threshold 설정을 통해 고객의 만족도를 최대화할 수 있는 가능성을 시사한다.

% Threshold가 변화함에 따라 latency의 감소 비율 및 성능 감소 비율. 

%\subsection{Hyper param에 따른 결과추이}
% 검색 모델, 

\section{Limitation}
\begin{table}[t]
  \centering
\scriptsize
\caption{Time and cost requirements for QA generation.}
\vspace{-4mm}
  \begin{tabular}{lcccccc}
    \toprule
    & \textbf{\# of QA}
    & \makecell[c]{\textbf{Layout}\\\textbf{analysis}}
    & \makecell[c]{\textbf{Description}\\\textbf{generation}}
    & \makecell[c]{\textbf{Chunk}\\\textbf{hierarchy}}
    & \makecell[c]{\textbf{QA}\\\textbf{generation}}
    & \makecell[c]{\textbf{API}\\\textbf{cost}}\\
\midrule
Law            & 16,134  & 237m  & 258m & 159m & 460m  & 2.91\$ \\
Finance        & 40,226  & 2,191m & 644m & 565m & 1,151m & 7.74\$ \\
Textbook       & 5,886   & 326m  & 247m & 26m  & 1,104m & 0.96\$ \\
Manual         & 17,477  & 531m  & 280m & 204m & 1,884m & 3.16\$ \\
Newspaper      & 24,931  & 220m  & 431m & 118m & 486m  & 4.25\$ \\
Academic       & 14,991  & 480m  & 245m & 152m & 448m  & 2.72\$ \\
Administ.       & 15,568  & 616m  & 249m & 162m & 905m  & 2.76\$ \\
\bottomrule
\end{tabular}
\label{tab:cost}
\vspace{-1mm}
\end{table}

Despite its benefits, HybridRAG incurs one-time offline overhead costs during the QA pre-generation. Generating the QA base of approximately 130,000 pairs demands computational resources and API expenses summarized in Table \ref{tab:cost}. However, these upfront costs can be justified given the substantial gains in query time performance and response quality. We believe these gains would be more valuable in real-world chatbot use cases which need to address large amounts of unstructured documents and numerous users with limited GPU resources.

%Future research could focus on optimizing pre-generation efficiency or developing incremental QA updating methods.

\section{Conclusion}
We introduced HybridRAG, a practical RAG framework optimized for real-world chatbot scenarios that need to provide low-latency and address unstructured scanned PDF documents. HybridRAG analyzes PDF documents via layout detection, OCR, hierarchical chunking, and dedicated LLM-based description generation for visual elements, and then generates a rich repository of plausible QA pairs. At query time, user queries are matched against the pre-generated QA pairs via embedding similarity; if a sufficiently similar question is found, the corresponding answer is directly returned; if not, an LLM generates responses based on the retrieved contexts. Validated through extensive experiments on OHRBench, HybridRAG achieves higher answer quality (with improved F1, ROUGE-L, and BERTScore) and lower latency compared to the standard RAG baseline.

\section*{Acknowledgement}
This work was supported by \textbf{Makebot Inc.} and the Institute of Information \& communications Technology Planning \& Evaluation(IITP) grant funded by the Korea government(MSIT)(\textbf{RS-2020-II201373}, Artificial Intelligence Graduate School Program (Hanyang University)).  

%\clearpage
%\newpage
\balance
\bibliographystyle{ACM-Reference-Format}
\bibliography{sample-base}

\end{document}